%% file: paper.tex
\documentclass[journal,twoside]{IEEEtran}
\usepackage{amsmath,amsfonts}
\usepackage{algorithmic}
\usepackage{algorithm}
\usepackage{array}
\usepackage[caption=false,font=scriptsize,labelfont=sf,textfont=sf]{subfig}
\usepackage{textcomp}
\usepackage{stfloats}
\usepackage{url}
\usepackage{verbatim}
\usepackage{graphicx}
\usepackage{cite}
\usepackage{multirow}
\usepackage{tabularx}
\usepackage{color}
\usepackage{hyperref}
\usepackage{bm}
\usepackage{booktabs}
\usepackage{siunitx}
\usepackage{threeparttable}
\usepackage{xcolor}
\usepackage{tcolorbox}
\tcbuselibrary{skins}
\tcbuselibrary{breakable}
\usepackage{tabularx}
\usepackage{colortbl}
\usepackage{fontawesome5}
\usepackage{enumitem} 

\hypersetup{
    colorlinks=true,
    linkcolor=blue,
    filecolor=blue,
    urlcolor=blue,
    citecolor=blue
}
\definecolor{lime}{HTML}{A6CE39}
\DeclareRobustCommand{\orcidicon}{%
	\begin{tikzpicture}
	\draw[lime, fill=lime] (0,0) 
	circle [radius=0.16] 
	node[white] {{\fontfamily{qag}\selectfont \tiny ID}};
	\draw[white, fill=white] (-0.0625,0.095) 
	circle [radius=0.007];
	\end{tikzpicture}
	\hspace{-3mm}
}

\foreach \x in {A, ..., Z}{%
	\expandafter\xdef\csname orcid\x\endcsname{\noexpand\href{https://orcid.org/\csname orcidauthor\x\endcsname}{\noexpand\orcidicon}}
}

\definecolor{HeaderGray}{gray}{0.95} 
\definecolor{LineGray}{gray}{0.6}   
\definecolor{Blue}{HTML}{0000CC}   
\definecolor{Green}{HTML}{00CC00} 
\definecolor{Purple}{HTML}{800080} 
\definecolor{Yellow}{HTML}{FFA500} 
\definecolor{Pink}{HTML}{FF00FF}   
\definecolor{mySlateGray}{HTML}{2F4F4F}

\definecolor{myLightGray}{rgb}{0.95, 0.95, 0.95} 
\definecolor{myDarkGray}{rgb}{0.2, 0.2, 0.2}   
\definecolor{myMediumGray}{rgb}{0.4, 0.4, 0.4} 
\definecolor{myFaintGray}{rgb}{0.6, 0.6, 0.6} 
\definecolor{myTitleColor}{rgb}{0.1, 0.1, 0.4} 

\definecolor{myLightBlueGrey}{rgb}{0.92, 0.95, 0.96} 
\definecolor{myDarkBlueGrey}{rgb}{0.25, 0.35, 0.45}   
\definecolor{myMediumBlueGrey}{rgb}{0.5, 0.6, 0.7} 
\definecolor{myFaintBlueGrey}{rgb}{0.7, 0.75, 0.8} 
\definecolor{mySectionTitleBlue}{rgb}{0.0, 0.0, 0.5} 
\definecolor{myGreen}{rgb}{0.0, 0.5, 0.0} 

\newcommand{\secondarykeyword}[1]{\textcolor{Pink}{#1}}
\newcommand{\keyword}[1]{\textcolor{myGreen}{\textbf{#1}}}

\newtcolorbox{examplebox}{
  freelance, 
  notitle,   
  boxrule=0.5pt, 
  colframe=black, 
  colback=HeaderGray, 
  arc=2pt,   
  outer arc=2pt, 
  boxsep=0pt,  
  left=5pt, right=5pt, top=5pt, bottom=5pt, 
}

\begin{document}

\title{DeltaVLM: Interactive Remote Sensing Image Change Analysis via Instruction-guided Difference Perception}

\author{Pei~Deng$^\dagger$, Wenqian~Zhou$^\dagger$,~\IEEEmembership{Student Member,~IEEE}, Hanlin~Wu\orcidA{},~\IEEEmembership{Member,~IEEE}
    \thanks{$^\dagger$The ﬁrst two authors contributed equally to this work. This work was supported in part by the National Natural Science Foundation of China under Grant 62401064, and in part by the Fundamental Research Funds for the Central Universities under Grant 2024TD001. \emph{(Corresponding author: Hanlin Wu.)}
    }
    \thanks{
    The authors are with the School of Information Science and Technology, Beijing Foreign Studies University, Beijing 100875, China.
    (e-mail: hlwu@bfsu.edu.cn).}
}

\markboth{}%
{}

\IEEEpubid{}

\maketitle

\begin{abstract}
    Accurate interpretation of land‐cover changes in multi‐temporal satellite imagery is critical for real-world scenarios. However, existing methods typically provide only one‐shot change masks or static captions, limiting their ability to support interactive, query‐driven analysis. In this work, we introduce remote sensing image change analysis (RSICA) as a new paradigm that combines the strengths of change detection and visual question answering to enable multi‐turn, instruction‐guided exploration of changes in bi-temporal remote sensing images. To support this task, we construct ChangeChat-105k, a large-scale instruction-following dataset, generated through a hybrid rule-based and GPT-assisted process, covering six interaction types: change captioning, classification, quantification, localization, open-ended question answering, and multi-turn dialogues. Building on this dataset, we propose DeltaVLM, an end-to-end architecture tailored for interactive RSICA. DeltaVLM features three innovations: (1) a fine-tuned bi-temporal vision encoder to capture temporal differences; (2) a visual difference perception module with a cross-semantic relation measuring (CSRM) mechanism to interpret changes; and (3) an instruction-guided Q-former to effectively extract query-relevant difference information from visual changes, aligning them with textual instructions. We train DeltaVLM on ChangeChat-105k using a frozen large language model, adapting only the vision and alignment modules to optimize efficiency. Extensive experiments and ablation studies demonstrate that DeltaVLM achieves state-of-the-art performance on both single‑turn captioning and multi‑turn interactive change analysis, outperforming existing multimodal large language models and remote sensing vision-language models. Code, dataset and pre-trained weights are available at \url{https://github.com/hanlinwu/DeltaVLM}.
\end{abstract}


\section{Introduction}
\label{sec:introduction}

\begin{figure}[!t]
    \centering
    \includegraphics[width=\linewidth]{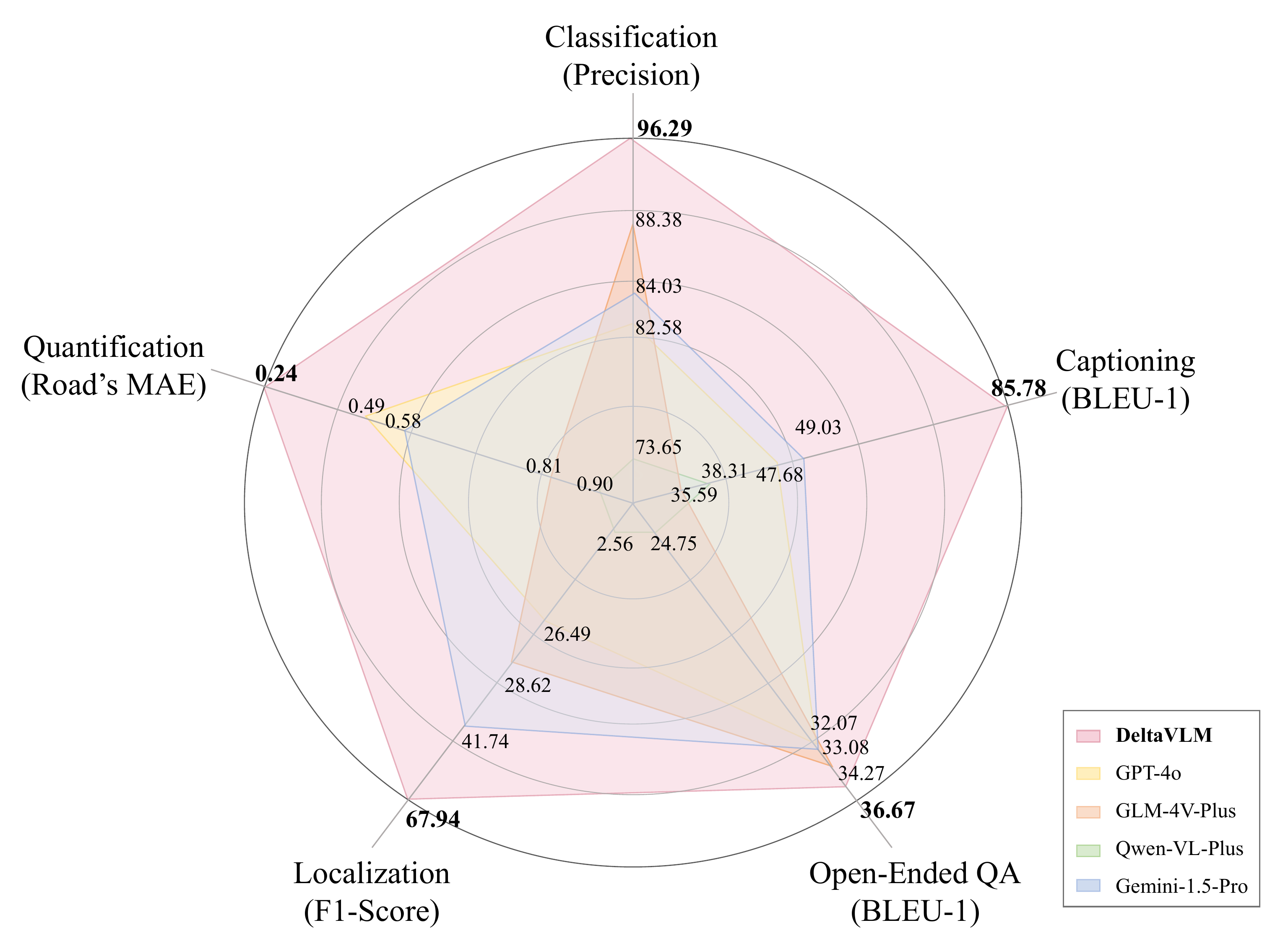}
    \caption{The performance of DeltaVLM against state‑of‑the‑art VLMs on five RS change analysis tasks. Each axis corresponds to a task‑specific metric: captioning (BLEU‑1), classification (precision), quantification (inverted Road's‑MAE), localization (F1‑score), and open‑ended QA (BLEU‑1).}
    \label{fig:radar}
\end{figure}

The continuous acquisition of vast amounts of data by Earth observation satellites opens up opportunities to monitor our dynamic planet through remote sensing images (RSIs). These images enable the extraction and interpretation of temporal changes, offering significant value for applications such as disaster management~\cite{van2000remote}, deforestation monitoring~\cite{chowdhury2006driving}, and environmental surveillance~\cite{navalgund2007remote}. However, analyzing RSIs poses distinct challenges compared to natural image processing, particularly in interpreting multi-temporal changes. Factors such as atmospheric variations, sensor differences, and geometric distortions~\cite{bannari1995theoretical} complicate the accurate detection and interpretation of changes over time. Early efforts to analyze temporal changes in RS imagery primarily relied on change detection techniques~\cite{ding2025survey}, such as pixel-level or object-based methods. While effective at locating changes, these methods often fail to provide insights into the nature of these changes.

Integrating natural language processing (NLP) techniques into RSIs interpretation has narrowed the gap between raw visual data and human understanding. Remote sensing image captioning (RSIC)~\cite{qu2016deep} aims to generate descriptive captions for a single observation. To enable more interactive exploration, remote sensing visual question answering (RSVQA)~\cite{lobry2020rsvqa} was introduced, allowing users to query RSIs with natural language questions and receive corresponding textual responses. However, RSIC and RSVQA are constrained to single-image analysis and lacks the ability to capture temporal changes. To address this limitation, RS image change captioning (RSICC)~\cite{liu2022remote} extends RSIC to bi-temporal analysis, generating textual descriptions of spatiotemporal differences.

More recently, the advent of large language models (LLMs) \cite{radford2019language} and their evolution into vision‑language models (VLMs)~\cite{Lu2019ViLBERTPT} have introduced interactive capabilities for RS interpretation, accommodating follow-up inquiries beyond static captions. However, these models, primarily trained on natural scenes, show limited performance on RS tasks due to significant data distribution differences, as illustrated in Fig.\,\ref{fig:radar}. To address this domain gap, recent work has adapted VLMs to RS tasks through instruction tuning on domain-specific datasets~\cite{wei2021finetuned}. This approach efficiently transfers VLMs' reasoning capabilities to RS interpretation, achieving strong performance on complex, open-ended tasks with relatively few training samples. For example, RSGPT~\cite{hu2025rsgpt} extends LLMs to RS tasks like RSIC and RSVQA via domain-adaptive fine-tuning. Similarly, models like GeoChat~\cite{kuckreja2024geochat} and RS-LLaVA~\cite{bazi2024rs}, built upon the LLaVA~\cite{liu2023visual} framework, excel in single-image region-based question answering (QA) and visual grounding. However, these models are limited to single-image analysis and cannot generate instruction-specific change descriptions from bi-temporal data. Moreover, most existing RS-VLMs primarily focus on fine-tuning LLM backbones, while overlooking RS-specific visual challenges in multi-temporal analysis, including atmospheric variations, seasonal changes, and sensor noise that can mask meaningful temporal differences~\cite{rasti2018noise}. This limitation is further compounded by the absence of large-scale instruction-following datasets for interactive bi-temporal RS analysis.

To address these challenges, we introduce remote sensing image change analysis (RSICA), a novel task that integrates the semantic grounding of change captioning with the interactive and reasoning capabilities of VQA. RSICA enables multi-turn, multi-task dialogues, allowing users to dynamically explore and interpret changes in bi-temporal RSIs. To support this task, we meticulously construct ChangeChat-105k, a large-scale instruction-following dataset with 105,107 instruction-response pairs, generated via a hybrid approach combining rule-based methods and ChatGPT's in-context learning capabilities~\cite{openai2022chatgpt}. Our proposed dataset covers a wide range of instruction types for interactive change analysis, including: 1) change captioning, 2) binary change detection, 3) category-specific change quantification, 4) change localization, 5) open-ended QA, and 6) multi-turn conversation. 

Building upon the ChangeChat-105k dataset, we present DeltaVLM, an innovative end-to-end architecture for query-driven RSICA. Unlike existing VLMs which target a single image as input, DeltaVLM extends the conventional three-stage change captioning pipeline into a specialized vision-language framework that enables instruction-guided, multi-task change interpretation of bi-temporal RSIs. Its core components include: 1) A bi-temporal vision encoder (Bi-VE), which processes bi-temporal RSIs to extract and compare features, capturing temporal differences at multiple scales; 2) an instruction-guided difference perception module (IDPM), incorporating a cross-semantic relation measuring (CSRM) mechanism and a Q-former to perceive subtle visual changes, filter out irrelevant semantics and noise in context, and dynamically align these differences with user-specific instructions; and 3) an LLM that decodes the aligned difference information into context-aware language responses.

To validate the effectiveness of DeltaVLM, we conduct comprehensive experiments and ablation studies on the proposed RSICA task using ChangeChat-105k. The results demonstrate that DeltaVLM achieves state-of-the-art (SOTA) performance, outperforming existing general-purpose large VLMs and RS change captioning models in interactive change analysis scenarios.

The main contributions of this paper can be summarized as follows:
\begin{enumerate}
    \item We introduce RSICA, a novel task that integrates change captioning and VQA into a unified, interactive, and user-driven framework for analyzing bi-temporal RSIs.
    \item We present ChangeChat-105k, a large-scale RS instruction-following dataset covering diverse change-related tasks covering captioning, classification, counting, localization, open-ended QA, and multi‑turn dialogues.
    \item We propose DeltaVLM, an innovative VLM architecture tailored for RSICA, which integrates a bi-temporal vision encoder, a visual difference perception module with CSRM, and a instruction-guided Q-former for dynamic, context-aware multi-task and multi-turn interactions.
    \item We conduct comprehensive evaluations, demonstrating DeltaVLM's superior performance compared to existing baselines on the RSICA task, validating its effectiveness in addressing complex change analysis challenges.
\end{enumerate}

\section{Related Work}

Our work on RSICA builds upon several key ideas including change detection, change captioning, VQA and VLMs. In this section, we provide a concise review of these areas.

\subsection{Change Detection}

Change detection in RS aims to identify and localize changes between satellite images over the same geographic region captured at different times. We categorize and review these methods into two main streams: traditional methods and deep learning (DL)-based methods.

\subsubsection{Traditional Change Detection Methods}

Early algebraic-based change detection methods, such as image differencing, ratioing, and change vector analysis~\cite{singh1989review}, were computationally simple but sensitive to noise and radiometric inconsistencies. To improve robustness, index-based~\cite{nelson1983detecting}, transformation-based~\cite{nielsen1998multivariate}, and classification-based methods~\cite{serra2003post} were developed, each addressing specific limitations but still lacking semantic understanding. Object-based image analysis~\cite{blaschke2010object} introduced spatial context, improving change detection accuracy in high-resolution RS imagery but depending heavily on accurate image segmentation. While these traditional methods contributed valuable concepts, they generally rely on handcrafted features and statistical models.

\subsubsection{DL-Based Change Detection Methods}

DL has transformed change detection in RS by enabling models to learn spatial and temporal representations directly from bi-temporal RSIs. Early methods based on CNNs, such as Siamese networks~\cite{daudt2018fully} and U-Net architectures~\cite{ronneberger2015u}, improved the accuracy of change localization through pairwise comparison or end-to-end segmentation. 

More recently, Transformer-based architectures~\cite{chen2021remote, bandara2022transformer} have demonstrated stronger performance by leveraging self-attention mechanisms to handle multi-scale features and model  temporal dependencies. They provide more global context compared to traditional CNNs. However, these models often require large-scale annotated datasets and high computational resources. To address the need for labeled data, self-supervised learning has gained traction, using strategies like contrastive learning~\cite{chen2020simple} and masked image modeling~\cite{bao2021beit}. For better generalization, few-shot and zero-shot learning methods~\cite{palatucci2009zero} have been developed. Futhermore, multimodal fusion has been applied to enhance detection stablity under adverse conditions like cloud-covered or rainy scenarios~\cite{longbotham2012multi}. A more comprehensive survey of change detection methods can be found in~\cite{ding2025survey}.

\subsection{Change Captioning}

While change detection primarily output binary masks, change captioning seeks to generate natural language descriptions of observed changes between bi-temporal RSIs.

\subsubsection{Encoder-Decoder-Based Methods}

Early change captioning methods primarily rely on encoder-decoder architectures: CNNs were used as visual encoders to extract features from  paired images, while RNNs served as language decoders to produce captions highlighting changes~\cite{hoxha2022change}. To improve caption quality, subsequent studies incorporated attention mechanisms into the decoder, enabling the model to dynamically focus on salient change regions~\cite{you2016image}. Although these methods laid the groundwork for change captioning, they often fail to effectively model global semantic relationships in complex RS scenes.

\subsubsection{Three-Stage-Based Methods}

Building upon early encoder–decoder frameworks, recent DL-based change captioning methods have evolved into a three-stage architecture: visual encoding, bi-temporal feature fusion, and language decoding. In the visual encoding stage, pretrained backbones such as ResNet or Vision Transformers (ViTs) are widely used to extract high-level features from the input imagery. Next, during temporal feature fusion, various task-specific strategies have been proposed to effectively integrate these features and emphasize change-related information. In the final language decoding stage, tansformer-based decoders have gradually replaced RNNs, owing to their superior global context modeling, which leads to generate more accurate and descriptive captions. Within this framework, several representative methods have emerged. Liu et al.~\cite{liu2022remote} introduced RSICCFormer, which employs ResNet-101 as the visual encoder and a dual-branch transformer decoder to enhance change representation. Similarly, Sun et al. proposed the sparse focus Transformer (SFT)~\cite{sun2024lightweight}, also leveraging ResNet-101 but incorporating a sparse attention mechanism to selectively focus on change regions while reducing computational cost. In contrast, ViT-based methods like PSNet~\cite{liu2023progressive} adopts ViT-B/32 as the encoder and optimizes its transformer decoder with multi-scale feature fusion, enabling more detailed and accurate change descriptions.

\subsubsection{Multi-Task and Semantic Alignment Methods}

While architectural improvements have advanced change captioning, they often fail to capture the intrinsic semantic relationships between change captioning and related tasks. To address this, recent methods have explored task-level innovations, particularly in multi-task learning and semantic alignment. These approaches often involve joint modeling of change captioning with auxiliary tasks, such as change detection, task decomposition, or semantic guidance to enhance the semantic quality of generated descriptions. For instance, Liu et al.\cite{liu2023decoupling} introduces PromptCC, which decouples the change captioning task into two subtasks: binary change classification and fine-grained change perception. This two-step process improves captioning performance compared to directly generating from raw bi-temporal RSIs. Building on this, Zhu et al. proposed Semantic-CC~\cite{Zhu2024SemanticCCBR}, which incorporates pixel-level semantic guidance from auxiliary change detection tasks to produce more accurate and contextually aligned captions. 

\subsubsection{LLM-Based Methods}

Although multi-task and semantic alignment methods have achieved notable progress, they are often tailored to specific tasks and datasets, lacking flexibility over diverse RS tasks. To address these limitations, recent research has turned to LLMs\cite{brown2020language}, leveraging their powerful generative and contextual reasoning abilities to enhance the robustness of change descriptions. Rather than training models from scratch, LLM-based approaches typically employ pre-trained vision encoders to extract high-level semantic features, which are then integrated with LLMs to generate context-aware captions. Two main strategies are commonly adopted to enhance the captioning quality: (1) fine-tuning LLMs on RS datasets with domain-specific annotations, and (2) guiding the captioning process with task-specific prompts or intermediate outputs (e.g., change maps) to help the model focus on relevant changes. For example, Noman et al.\cite{noman2024cdchat} developed CDChat, which fine-tunes Vicuna-v1.5 on datasets like SYSU-CD\cite{shi2021deeply} and LEVIR-CD~\cite{chen2020spatial}, aligning visual and textual representations through supervised learning. This leads to improved captioning accuracy and interpretability. Overall, LLM-based methods represent a promising new direction in change captioning. Their ability to generalize across tasks and adapt to diverse RS scenarios holds great potential for future research and real-world applications.

\subsection{RSVQA}

While change detection and change captioning have significantly advanced RS image understanding, they remain constrained by predefined information extraction and are typically limited to single-turn interactions and static outputs. To bridge this gap and enable more flexible, user-driven analysis, researchers have drawn inspiration from VQA~\cite{antol2015vqa} in the natural image domain to develop RSVQA. RSVQA\cite{lobry2020rsvqa} allows users to pose natural language questions tailored to specific analytical needs, with the VQA system generating contextually relevant answers based on the visual content of RSIs. Lobry et al.\cite{lobry2020rsvqa} established two benchmark datasets: RSVQA-LR and RSVQA-HR, comprising large-scale image-question-answer triplets from agricultural and urban scenes and covering low and high spatial resolutions, lay the groundwork for future research into more dynamic and user-centric RS applications.

Most RSVQA methods follow an encoder-fusion-decoder architecture. To improve fine-grained multimodal reasoning, several approaches have been proposed to enhance the fusion module. For instance, the spatial hierarchical reasoning network~\cite{zhang2023spatial} incorporates hash-based multi-scale attention mechanisms and semantic segmentation priors to better capture spatial hierarchies, while semantic object-awareness methods~\cite{wang2024earthvqa} strengthen spatial reasoning through object-level relational modeling. In contrast, Chappuis et al.~\cite{chappuis2022prompt} proposes Prompt-RSVQA, which offers an alternative to bi-modal fusion by translating visual information into textual prompts, with fusion occurring implicitly within a transformer-based language model. RSVQA methods have evolved from early-stage global feature fusion to advanced attention-based architectures, incorporating object-centric reasoning and transformer-based interactions.

Despite these advances in static image analysis, existing RSVQA methods often overlook the temporal dynamics inherent in geospatial change analysis of RS data\cite{liu2024remote}. Change-aware VQA~\cite{yuan2022change} was an early attempt at addressing this issue, proposing a change-aware visual question answering method based on multi-temporal aerial images. However, it employed a classifier to select answers rather than generating natural language, significantly reducing response flexibility compared to VLMs.

\subsection{VLMs}
VLMs \cite{li2022blip,alayrac2022flamingo,liu2023visual} combine visual encoders with LLMs to perform multimodal tasks, such as image captioning, cross-modal retrieval, and VQA. Unlike traditional VQA systems, which rely on task-specific architectures with separate visual and language modules, VLMs leverage a unified Transformer architecture with large-scale pretraining for versatile multimodal tasks. Advancements in LLMs' language understanding and reasoning capabilities have also driven the development of VLMs. Notable examples include GPT-4o\cite{hurst2024gpt}, Qwen-VL-Plus\cite{wang2024qwen2}, GLM-4V-Plus\cite{glm2024chatglm} and Gemini-1.5-Pro\cite{Reid2024Gemini1U}. Although these general VLMs perform well in zero-shot reasoning, they struggle in the RS domain due to different domain-specific semantics and challenges such as atmospheric noise and scale variations.

To address the challenges of applying VLMs in the RS domain, some researchers have constructed large-scale visual-language datasets \cite{zhang2024rs5m} specific to RS, fine-tuning pretrained VLMs and achieving notable progress. For instance, Kuckreja et al.~\cite{kuckreja2024geochat} developed GeoChat, a pioneering model that enables region-level conversational QA. Similarly, RSGPT~\cite{hu2025rsgpt} builds upon InstructBLIP~\cite{dai2023instructblip} with an annotation-aware Q-former module, which maps region-level visual features to instruction-conditioned query embeddings and improves visual–textual alignment for region-level conversational QA. Extending this, Zhan et al.~\cite{zhan2025skyeyegpt} propose SkyEyeGPT, further aligning spatially localized visual features with instruction-following LLMs. SkyEyeGPT establishes a multi-task conversational framework including segmentation, detection, grounding, and conversational QA. 

Although VLMs in the RS field have made great progress, they are designed for single-temporal analysis and cannot provide user-interactive analysis of dynamic changes in bi-temporal RSIs. Recently, ChangeChat~\cite{deng2025changechat} developed an instruction-following dataset and fine-tuned pretrained VLMs for temporally-aware geospatial tasks. However, ChangeChat did not differentially extract relevant visual features based on varying instructions, leaving room for further improvement in response quality. To address this issue, we expanded the ChangeChat dataset and redesigned DeltaVLM to extract instruction-guided differential information, enabling multi-task, multi-turn dialogues for user-driven interpretation of geospatial changes.

\input{imgs/instruction_types}

\section{ChangeChat-105k dataset}
Recent advances in RS change detection and captioning have produced benchmark datasets like LEVIR-CC~\cite{liu2022remote} and LEVIR-MCI~\cite{liu2024change}. LEVIR-CC provides 10,077 bitemporal image pairs with five human-written captions each, supporting change captioning but lacking fine-grained annotations for object counts or precise locations. LEVIR-MCI, built upon LEVIR-CC, provides pixel-level change maps to support binary change detection but lacks deeper exploration of change information and cannot support interactive analysis.

To support RSICA, we introduce ChangeChat-105k, a large-scale dataset with 105,107 instruction–response pairs derived from LEVIR-CC and LEVIR-MCI. We employ a hybrid pipeline combining rule-based methods and LLM-based generation, leveraging ChatGPT’s~\cite{openai2022chatgpt} in-context learning. For structured tasks like object counting and localization, we use rule-based methods with LEVIR-MCI’s pixel-level change maps and OpenCV-based contour detection to extract precise information. For open-ended tasks, we generate diverse instructions using ChatGPT with curated prompts and seed examples. The dataset includes six instruction types, ranging from structured information extraction to open-ended reasoning, as detailed in Fig.~\ref{tab:instruction_types}, enabling comprehensive evaluation of multi-task, interactive change analysis capabilities.

\input{imgs/data-pipeline}

\subsubsection{Change Captioning} In this task, we extend the original LEVIR-CC triplets ($I_{t_1}$, $I_{t_2}$, $C$) into an instruction-response format. For each bi-temporal image pair ($I_{t_1}$, $I_{t_2}$) and its corresponding change caption ($C$), we design a fixed instruction $Q$ as: ``\textit{Please briefly describe the changes in these two images.}'' The instruction-response pair is formatted as follows:

\begin{equation*}
    \begin{aligned}
        &\texttt{Human:}\quad I_{t_1}\quad I_{t_2}\quad Q \quad \texttt{<STOP>}\\
        &\texttt{Assistant:} \quad C \quad \texttt{<STOP>}.\\
    \end{aligned}
\end{equation*}

\subsubsection{Binary Change Classification} In this task, we generate instructions that ask DeltaVLM to determine whether changes occurred, expecting a binary ``\textit{yes}'' or ``\textit{no}'' answer. The instructions are designed with the template: ``\textit{Please judge whether these two images have changed or not. Answer yes or no.}'' The ground truth for each image pair is derived from the change map in LEVIR-MCI.

\subsubsection{Category-Specific Change Quantification} We created instructions to guide DeltaVLM in quantifying changes in specific categories, such as calculating the number of newly added buildings or roads. These quantity-related instructions are generated based on templates and use the OpenCV library's contour detector to calculate the number of objects.

\subsubsection{Change Localization}
To localize changes spatially, we design instructions that ask DeltaVLM to return changed regions in a $3\times3$ grid, with cells denoted as:
\[
P = \{%
\mathrm{TL},\,\mathrm{TC},\,\mathrm{TR},\,
\mathrm{CL},\,\mathrm{CC},\,\mathrm{CR},\,
\mathrm{BL},\,\mathrm{BC},\,\mathrm{BR}\},
\]
where TL=top-left, TC=top-center, …, BR=bottom-right. The ground truth is obtained by splitting each change map into nine blocks; any block with $>$5\% changed pixels is labeled as changed.

\subsubsection{Open-ended QA} To generate more diverse instruction-following data, we leverage ChatGPT's in-context learning capabilities to automatically generate instruction–response pairs. As shown in Fig.\,\ref{fig:data_generation_pipeline} We began by providing ChatGPT a system message to guide its responses. Then, we manually designed a few seed examples for each type of task to help it understand the desired output structure. Specifically, it generated two types of conversational data: i) Q-A pairs from change captions; and ii) fine-grained queries incorporating extracted contour and quantification information. 

Notably, we did not provide any visual information to ChatGPT. All questions and answers were derived from prompts we constructed based on five captions, as well as the change contours and counting information extracted from the change map, as illustrated in Fig.\,\ref{fig:data_generation_pipeline}\,(b).

\subsubsection{Multi-turn Conversation} We designed multi-turn dialogues to encourage DeltaVLM to perform change analysis using a chain-of-thought (CoT) approach. The instructions are presented in increasing difficulty, beginning with simple binary change classification, followed by change object and quantity identification, and progressing to the complex and detailed change captioning task.

\section{Methodology}
\label{sec:methodology}
\IEEEpubidadjcol
In this section, we provide a detailed explanation of the architecture of DeltaVLM.

\subsection{Overview}

\begin{figure*}[htbp]
    \centering
    \includegraphics[width=\linewidth]{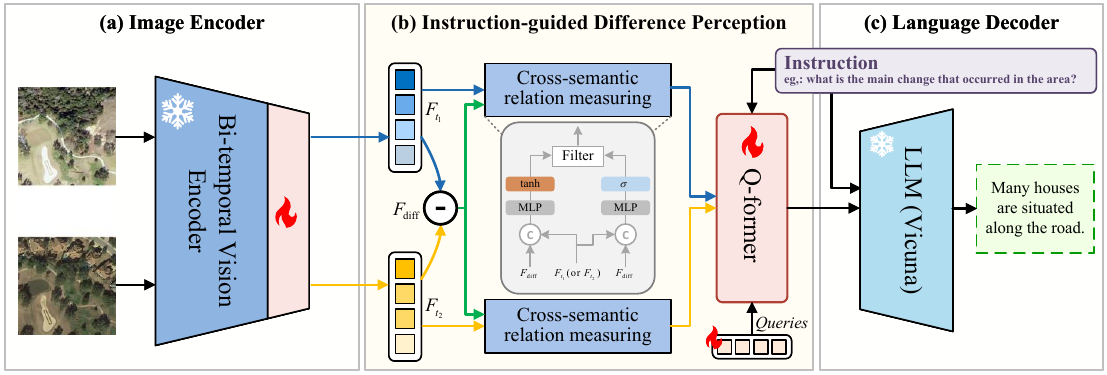}
    \caption{An overview of our proposed DeltaVLM.}
    \label{architecture}
\end{figure*}

As shown in Fig.\,\ref{architecture}, DeltaVLM is an end-to-end framework tailored for interactive RSICA, comprising three key steps: (1) bi-temporal visual feature encoding, (2) instruction-guided difference feature extraction, and (3) language decoding based on an LLM.

First, the bi-temporal vision encoder (Bi-VE) extract features from the paired input images $I_{t_1}$ and $I_{t_2}$, 
\begin{equation}
    F_{t_1}, F_{t_2} = \Phi_{\text{Bi-VE}}(I_{t_1}, I_{t_2}).
\end{equation}

Then, the IDPM enhances the bi-temporal features $F_{t_1}, F_{t_2}$ with a CSRM mechanism, followed by a Q-former that aligns the refined features with user instruction $P$ and learnable queries $Q$, producing instruction-guided difference representations $\hat{F}_{\text{diff}}$,
\begin{equation}
    \begin{aligned}
        F'_{t_1}, F'_{t_2} &= \Phi_{\text{enhancer}}(F_{t_1}, F_{t_2}) \\
        \hat{F}_{\text{diff}} &= \Phi_{\text{Q-former}}([F'_{t_1}, F'_{t_2}]; P, Q).
    \end{aligned}
\end{equation}

Finally, $\hat{F}_{\text{diff}}$ are decoded by an LLM into a natural language response $T$, conditioned on the instuction $P$,
\begin{equation}
    T = \Phi_{\text{LLM}}(\hat{F}_{\text{diff}}, P).
\end{equation}

\subsection{Bi-temporal Vision Encoding}
\label{sec:bive}

To leverage the power of large-scale pretraining for our Bi-VE, we adopt the EVA-ViT-g/14~\cite{fang2023eva} as its backbone. To adapt it to RSICA while mitigating catastrophic forgetting, we employ selective fine-tuning: the first 37 Transformer layers are frozen, and only the final two blocks are fine-tuned.

Given a bi-temporal RS image pair $I_{t_1}, I_{t_2} \in \mathbb{R}^{H \times W \times 3}$, where $H$ and $W$ denote the height and width, respectively, $\Phi_{\text{Bi-VE}}$ processes each image independently to avoid early fusion of temporal information and prevent biases in initial feature extraction. Each image is first divided into a sequence of $16 \times 16$ patch embeddings, which are then passed through the Transformer encoder layers to capture complex visual patterns. Features are extracted from the second-to-last layer (bypassing the classification head) for task-specific semantics, yielding $F_{t_1}, F_{t_2} \in \mathbb{R}^{N \times D}$, where $N$ is the number of patches and $D$ is the hidden dimension.

Mathematically, the Bi-VE's operation is given by:
\begin{equation}
    \begin{aligned}
    F_{t_1} &= \Phi_{\text{ViT}}(I_{t_1}; \Theta_{\text{fine-tuned}}) \in \mathbb{R}^{\frac{H}{16} \times \frac{W}{16} \times D}  \\
    F_{t_2} &= \Phi_{\text{ViT}}(I_{t_2}; \Theta_{\text{fine-tuned}}) \in \mathbb{R}^{\frac{H}{16} \times \frac{W}{16} \times D},
\end{aligned}
\end{equation}
where $\Phi_{\text{ViT}}$ is the EVA-ViT-g/14 encoder, $\Theta_{\text{fine-tuned}}$ denotes the parameters of the fine-tuned layers. The spatial resolution of $F_{t_1}$ and $F_{t_2}$ is reduced by a factor of 16 due to patch-based processing. These feature maps are then fed into the IDPM.

\subsection{Instruction-guided Difference Perception} 

Given bi-temporal features $F_{t_1}, F_{t_2}$, we first computes the raw visual difference,
\begin{equation}
    F_{\text{diff}} = F_{t_2} - F_{t_1},
\end{equation}
where $F_{\text{diff}} \in \mathbb{R}^{N \times D}$ captures all pixel-level changes. Directly decoding $F_{\text{diff}}$ into language may introduce interference information such as sensor differences, lighting, or seasonal variations. To address this problem, we first explore the semantic relationships between $F_{\text{diff}} $, $F_{t_1}$, and $F_{t_2}$ through a CSRM mechanism to eliminate irrelevant change interferences. Subsequently, cross-modal alignment is achieved through an instruction-guided Q-former.

\subsubsection{Cross-Semantic Relation Measuring} 

The CSRM machanism works in three steps: contextualizing, gating and filtering.

\textbf{Contextualizing.}  To understand how changes relate to each temporal state, we compute context vectors by fusing difference features with original features,
\begin{equation}
    \begin{aligned}
        C_{t_1} &= \tanh(W_c[F_{\text{diff}}; F_{t_1}] + b_c) \\
        C_{t_2} &= \tanh(W'_c[F_{\text{diff}}; F_{t_2}] + b'_c),
    \end{aligned}
\end{equation}
where $[\cdot\,; \cdot]$ denotes channel-wise concatenation, $W_c, W'_c \in \mathbb{R}^{D \times 2D}$ and $b_c, b'_c \in \mathbb{R}^{D}$ are learnable weights and biases. These linear projections transform the concatenated features into a new space that emphasizes semantic connections, while the $\tanh$ activation constrains the outputs to $[-1, 1]$.

\textbf{Gating.} Through contextualizing, these context vectors capture change-context relationships. To further weight each detected change by its semantic relevance, we then employ a gating mechanism, inspired by gated recurrent units (GRUs)~\cite{Cho2014LearningPR}. This step generates gate vectors $G_{t_1}$ and $G_{t_2}$ via a sigmoid activation:
\begin{equation}
    \begin{aligned}
        G_{t_1} &= \sigma(W_{\text{g}}[F_{\text{diff}}; F_{t_1}] + b_{\text{g}})\\
        G_{t_2} &= \sigma(W'_{\text{g}}[F_{\text{diff}}; F_{t_2}] + b'_{\text{g}}),
    \end{aligned}
\end{equation}
where $\sigma$ is the sigmoid function producing values in $(0, 1)$ as relevance scores, $W_g, W'_g \in \mathbb{R}^{D \times 2D}$ and $b_g, b'_g \in \mathbb{R}^{D}$ are learnable weights and biases.

\textbf{Filtering.} Finally, we selectively retain semantically relevant information through element-wise multiplication ($\odot$) between gate vectors and and their corresponding context vectors:
\begin{equation}
    \begin{aligned}
        F'_{t_1} &= G_{t_1} \odot C_{t_1} \\
        F'_{t_2} &= G_{t_2} \odot C_{t_2}.
    \end{aligned}
\end{equation}

This multiplication refines $C_{t_1}$ and $C_{t_2}$ under the guidance of $G_{t_1}$ and $G_{t_2}$ by suppressing irrelevant components (e.g., noise) with low gate values while preserving important changes (e.g., new structures or land cover shifts) with high values. Consequently, the resulting filtered features $F'_{t_1}$ and $F'_{t_2}$ retain only semantically relevant changes, which are then passed to the subsequent Q-former module.

\subsubsection{Q-former for Cross-Modal Alignment}

Inspired by InstructBLIP~\cite{dai2023instructblip}, our Q-former module is specifically designed to generate change-aware features aligned with the given instruction $P$. This process begins with a set of learnable query embeddings $Q \in \mathbb{R}^{L \times d}$, where $L=32$ is the number of queries and $d$ is the feature dimension matching the LLM's input space. These queries are first refined through self-attention layers:
\begin{equation}
    Q_{\text{SA}} = \text{SelfAttention}(Q). 
\end{equation}
Next, the refined queries $Q_{\text{SA}}$ attend to both the concatenated visual features and the instruction prompt through cross-attention:
\begin{equation}
    {Q}_{\text{CA}} = \text{CrossAttention}(Q_{\text{SA}}, [F'_{t_1}; F'_{t_2}], P).
\end{equation}
This step dynamically aligns the change features with the task-specific instruction, tailoring them to the user's query. Finally, the instruction-aware change features pass through a feed-forward network to yield the final compact output:
\begin{equation}
    \hat{F}_{\text{diff}} = \text{FFN}(Q_{\text{CA}}) \in \mathbb{R}^{32 \times d}. 
\end{equation}
By incorporating these steps, the Q-former ensures that the extracted features effectively capture instruction-relevant changes while maintaining computational efficiency through the query bottleneck.

\subsection{LLM-based Language Decoder}
We choose Vicuna-7B~\cite{zheng2023judging} as our language decoder, which is a powerful decoder-only LLM fine-tuned from LLaMA~\cite{touvron2023llama}. Our decoder takes the visual features $\hat{F}_{\text{diff}}$ and instruction prompt $P$ as input, generating instruction-specific change descriptions. The process begins by tokenizing and embedding the instruction prompt $P$.
\begin{equation}
    E = \Phi_{\text{embedding}}(P), \label{eq:tokenizer}
\end{equation}
where $\Phi_{\text{embedding}}$ denotes the tokenizer and embedding function, transforming the raw text into a sequence of embeddings $E$ suitable for the language model. These embeddings capture the semantic essence of the prompt by mapping words or sub-words to high-dimensional vectors.

The learned prompt-aligned features $\hat{F}_{\text{diff}}$, along with the embedded prompt $E$, serve as input to the language decoder, which generates a descriptive caption $T = \{t_1, \dots, t_N\}$ that summarizes the bi-temporal changes:
\begin{equation}
    T = \Phi_{\text{LLM}}(\hat{F}_{\text{diff}}, E) \in \mathcal{C}^N,
\end{equation}
where $\Phi_{\text{LLM}}$ denotes the Vicuna-7B decoder, and $T$ is a sequence of $N$ tokens from the vocabulary $\mathcal{C}$. This process effectively interprets the visual differences in the context of the user's query, generating task-specific change descriptions.

\subsection{Training Objective}

Unlike typical RSICC methods that generate fixed descriptions for RS image pairs, DeltaVLM is trained on instruction-conditioned data. To enable this, we augment the dataset with instruction prompts $P_j$, creating $D_{\text{train}} = \{(I_1, P_1, T_1), \cdots, (I_M, P_M, T_M)\}$, where each $P_j$ is a user query corresponding to the bi-temporal image pair $I_j$ and its target description $T_j$, allowing the model to adapt its output to diverse user instructions. The model is trained using the cross-entropy loss function:
\begin{equation}
    \mathcal{L}_{\text{train}} = -\frac{1}{K} \sum_{i=1}^{K} w_i \log(\hat{w}_i),
\end{equation}
where $K$ represents the total number of tokens in the target description, $w_i$ is the one-hot encoded ground-truth token at position $i$, and $\hat{w}_i$ is DeltaVLM's predicted probability for the $i$-th token. By minimizing this loss over the augmented dataset $D'_{\text{train}}$, we train DeltaVLM to generate accurate and contextually relevant descriptions tailored to user-specific instructions.

\section{Experiments and Analysis}
\label{sec:experiments}

In this section, we present comprehensive experiments to evaluate the effectiveness of DeltaVLM for RSICA. We first describe the experimental setup, then report quantitative results across multiple tasks, and finally analyze key components through ablation studies.

\subsection{Experimental Setup}

\subsubsection{Datasets} We evaluate our proposed DeltaVLM on the ChangeChat-105k dataset, which comprises 105,107 instruction–response pairs aligned with bi-temporal image patches of size $256 \times 256$ at $0.5\,\si{m/pixel}$ spatial resolution. Each image pair is annotated with multiple task types: binary change detection, object counting, change localization, and change captioning. For evaluation, we split the dataset into a training set and a test set, with the detailed distribution of instruction–response pairs across tasks and subsets provided in Table~\ref{tab:overview-dataset}.

\begin{table*}[htbp]
    \centering
    \setlength{\tabcolsep}{7pt}
    \renewcommand{\arraystretch}{1.1}
    \caption{Overview of the ChangeChat-105k Dataset: Instruction Types, Generation Methods, and Training/Testing Splits}
    \label{tab:overview-dataset}
    \begin{tabular}{l|c|c|c|c|c}
        \toprule[1.5pt]
        \textbf{Instruction Type} & \textbf{Source Data} & \textbf{Generation Method} & \textbf{Response Format} & \textbf{Training Set} & \textbf{Test Set} \\ 
        \midrule\midrule
        Change Captioning & LEVIR-CC & Rule-based & Descriptive Text & 34,075 & 1,929 \\
        Binary Change Classification & LEVIR-MCI & Rule-based & Yes/No Response & 6,815 & 1,929 \\
        Category-specific Change Quantification & LEVIR-MCI & Rule-based & Object Count & 6,815 & 1,929 \\
        Change Localization & LEVIR-MCI & Rule-based & Grid Location & 6,815 & 1,929 \\
        Open-ended QA & Derived (LEVIR-CC/MCI) & GPT-assisted & Q\&A Pair & 26,600 & 7,527 \\
        Multi-turn Conversation & Derived (LEVIR-MCI) & Rule-based & Multi-turn Dialogue & 6,815 & 1,929 \\
        \midrule\midrule
        \textbf{Total} & -- & -- & -- & \textbf{87,935} & \textbf{17,172} \\
        \bottomrule[1.5pt]
    \end{tabular}
\end{table*}

\subsubsection{Implementation Details} All experiments were conducted on Ubuntu 20.04 using the PyTorch framework on NVIDIA L20 GPUs. For data augmentation, we first apply random cropping that removes $0-5\%$ of the image content, followed by random rotation within $[-15^\circ, +15^\circ]$. The augmented images are then resized to $224 \times 224$ pixels to match the ViT-g/14 backbone's patch embedding requirements. We employed the AdamW optimizer~\cite{loshchilov2017decoupled} with a weight decay of 0.05 for regularization. The initial learning rate was set to $1 \times 10^{-5}$, with a batch size of 24 and a maximum of 30 training epochs. 


\subsubsection{Evaluation Metrics}

To comprehensively evaluate DeltaVLM's multi-task capabilities, we employ task-specific metrics that are widely adopted in the field to provide a holistic view of DeltaVLM's effectiveness.

\begin{itemize}
    \item {Change Captioning}: We adopt BLEU-N (N=1, 2, 3, 4)~\cite{papineni2002bleu}, METEOR~\cite{banerjee2005meteor}, ROUGE-L~\cite{lin2004rouge}, and CIDEr~\cite{Vedantam2014CIDErCI} to assess the quality of generated change descriptions. These metrics evaluate n-gram overlap, semantic similarity, sentence structure, and human consensus alignment, respectively.

    \item {Binary Change Classification}: We use accuracy, precision, recall, and F1-score to measure classification performance. The F1-score provide a balanced measure particularly important for change/no-change distributions.

    \item {Category-Specific Change Quantification}:  We use mean absolute error (MAE) and root mean squared error (RMSE) to evaluate counting accuracy, with MAE capturing average deviation and RMSE penalizing larger errors.

    \item {Change Localization}: The change localization task requires returning the location of the change in a $3\times 3$ grid format, and this task belongs to multi-class classification tasks. We use precision, recall, F1-score, Jaccard similarity, and subset accuracy to evaluate localization quality.
\end{itemize}

\subsection{Comparison with Baselines}

We compare DeltaVLM against SOTA baselines across various tasks: change captioning, binary change classification, category-specific change quantification, and change localization. The baselines are categorized into two groups: (1) RS-specific change captioning models, which focus on domain-adapted architectures for RS imagery, including RSICCFormer~\cite{liu2022remote}, Prompt-CC~\cite{liu2023decoupling}, PSNet~\cite{liu2023progressive}, and SFT~\cite{sun2024lightweight}; and (2) general-purpose large VLMs, including GPT-4o~\cite{hurst2024gpt}, Qwen-VL-Plus~\cite{wang2024qwen2}, GLM-4V-Plus~\cite{glm2024chatglm}, and Gemini-1.5-Pro~\cite{Reid2024Gemini1U}.

\subsubsection{Change Captioning} We first evaluate DeltaVLM against both specialized RS change captioning and general-purpose VLMs baselines on the ChangeChat-105k test set. As shown in Table\,\ref{tab:comparison_captioning}, we achieves SOTA performance in most metrics, demonstrating the effectiveness of instruction-guided change analysis. Among general-purpose VLMs, all models show significantly lower performance across all metrics, with BLEU-4 scores ranging from 13.85 (GLM-4V-Plus) to 22.95 (Qwen-VL-Plus), compared to over 62 for specialized models. This substantial gap highlights the challenge of applying general VLMs to RS change analysis without domain-specific fine-tuning, as they struggle to capture nuanced details like land cover shifts or structural changes in satellite imagery. In contrast, RS-specific models demonstrate stronger performance, leveraging architectures optimized for RS tasks. While SFT edges out in BLEU-4, METEOR, and CIDEr, likely due to its optimization for n-gram consensus during fine-tuning. DeltaVLM's superior performance in early n-gram metrics indicates more accurate word-level and phrase-level matching, suggesting our instruction-guided approach generates more contextually relevant change descriptions aligned with user queries. 

\begin{table*}[htbp]
    \centering
    \caption{Comparison with SOTA methods on Change Captioning Task on the ChangeChat-105k Dataset}
    \label{tab:comparison_captioning}
    \setlength{\tabcolsep}{7.5pt}
    \begin{tabular}{l|l|ccccccc}
        \toprule[1.5pt]
        Category & Method & BLEU-1 & BLEU-2 & BLEU-3 & BLEU-4 & METEOR & ROUGE-L & CIDEr-D \\
        \midrule\midrule
        \multirow{4}{*}{VLMs}
        & GPT-4o\cite{hurst2024gpt}         & 46.03 & 33.09 & 24.66 & 18.05 & 22.50 & 56.49 & 90.92 \\
        & Qwen-VL-Plus\cite{wang2024qwen2}    & 41.31 & 33.19 & 27.96 & 22.95 & 18.04 & 51.24 & 92.99 \\
        & GLM-4V-Plus\cite{glm2024chatglm}      & 35.59 & 24.26 & 18.54 & 13.85 & 20.13 & 54.39 & 93.16 \\
        & Gemini-1.5-Pro\cite{Reid2024Gemini1U}   & 45.68 & 33.59 & 25.53 & 19.01 & 22.64 & 56.25 & 91.37 \\
        \midrule
        \multirow{5}{*}{RS Change Captioning Models}
        & RSICCFormer\cite{liu2022remote}         & 84.72 & 76.27 & 68.87 & 62.77 & 39.61 & 74.12 & 134.12 \\
        & PromptCC\cite{liu2023decoupling}         & 83.66 & 75.73 & 69.10 & 63.54 & 38.82 & 73.72 & 136.44 \\
        & PSNet\cite{liu2023progressive}          & 83.86 & 75.13 & 67.89 & 62.11 & 38.80 & 73.60 & 132.62 \\
        & SFT\cite{sun2024lightweight}                     & 84.56 & 75.87 & 68.64 & \textbf{62.87} & \textbf{39.93} & 74.69 & \textbf{137.05} \\
        & \textbf{Ours}                           & \textbf{85.78} & \textbf{77.15} & \textbf{69.24} & 62.51 & 39.47 & \textbf{75.01} & 136.72 \\
        \bottomrule[1.5pt]
    \end{tabular}%
\end{table*}

\subsubsection{Binary Change Classification} For binary change classification, which determines whether changes occurred between bi-temporal RSIs based on user instructions, we compare DeltaVLM against large VLMs and achieve superior performance across all metrics, with an accuracy of $93.99\%$, precision of $96.29\%$, recall of $91.49\%$, and F1-score of $93.83\%$. Among the baseline models, performance also varies considerably. GPT-4o and Gemini-1.5-Pro demonstrate competitive results with F1-scores of $85.07\%$ and $83.77\%$ respectively, showing their capability in basic change detection tasks. However, models like Qwen-VL-Plus lag significantly with only 25.52\% recall and 37.90\% F1-score, indicating a strong bias toward predicting ``no change'' cases. Furthermore, DeltaVLM's superior performance represents improvements of $8.76\%$ in F1-score over the best baseline (GPT-4o), with particularly strong gains in precision ($+7.91\%$) and recall ($+4.87\%$). The balanced precision-recall trade-off ($96.29\%$ vs. $91.49\%$) demonstrates that our model enhanced capability in both correctly identifying changed regions and minimizing false positives.

\begin{table}[t]
  \centering
  \caption{Results on Binary Change Classification}
  \label{tab:comparison_classification}
  \setlength{\tabcolsep}{3pt}
  \renewcommand{\arraystretch}{1.1}
  \begin{tabular}{lcccc}
    \toprule[1.5pt]
    Method & Accuracy (\%) & Precision (\%) & Recall (\%) & F1 (\%) \\
    \midrule\midrule
    GPT-4o\cite{hurst2024gpt}       & 84.81 & 83.58 & 86.62 & 85.07 \\
    Qwen-VL-Plus\cite{wang2024qwen2}  & 58.22 & 73.65 & 25.52 & 37.90 \\
    GLM-4V-Plus\cite{glm2024chatglm}    & 79.83 & 88.38 & 68.67 & 77.29 \\
    Gemini-1.5-Pro\cite{Reid2024Gemini1U} & 83.83 & 84.03 & 83.51 & 83.77 \\
    \textbf{Ours}                                & \textbf{93.99} & \textbf{96.29} & \textbf{91.49} & \textbf{93.83} \\
    \bottomrule[1.5pt]
  \end{tabular}
\end{table}

\subsubsection{Category-Specific Change Quantification} Beyond simply detecting change, we assessed the models' ability to count specific object changes in two critical infrastructure categories: roads and buildings, using MAE and RMSE metrics to measure counting accuracy. As shown in Table \ref{tab:comparison_quantification}, DeltaVLM achieves the lowest errors across both categories, with MAE/RMSE of $0.24/0.70$ for roads and $1.32/2.89$ for buildings. For road counting, our model reduces MAE by $51\%$ compared to the best-performing baseline, GPT-4o, demonstrating precise detection of linear infrastructure changes. The low RMSE indicates consistent performance without large counting errors. Building quantification proves more challenging for all models due to varying sizes and occlusion patterns, yet DeltaVLM still achieves a $29\%$ MAE reduction over GPT-4o. DeltaVLM also achieves averaging $35\%$'s improvement across all metrics, which demonstrates our CSRM module's effectiveness in focusing on relevant object-specific changes while filtering out irrelevant variations.

\begin{table}[t]
  \centering
  \caption{Results on Change Quantification}
  \label{tab:comparison_quantification}
  \setlength{\tabcolsep}{9pt}
  \renewcommand{\arraystretch}{1.1}
  \begin{tabular}{lcccc}  
    \toprule[1.5pt]
    \multirow{2}{*}[-1ex]{Method} & \multicolumn{2}{c}{Roads} & \multicolumn{2}{c}{Buildings} \\
    \cmidrule(lr){2-3} \cmidrule(lr){4-5}
    & MAE & RMSE & MAE & RMSE \\
    \midrule\midrule
    GPT-4o\cite{hurst2024gpt}       & 0.49 & 1.00 & 1.86 & 4.57 \\
    Qwen-VL-Plus\cite{wang2024qwen2}  & 0.90 & 1.50 & 4.41 & 9.03 \\
    GLM-4V-Plus\cite{glm2024chatglm}    & 0.82 & 1.62 & 2.05 & 4.61 \\
    Gemini-1.5-Pro\cite{Reid2024Gemini1U} & 0.58 & 1.25 & 2.56 & 8.71 \\
    \textbf{Ours}                               & \textbf{0.24} & \textbf{0.70} & \textbf{1.32} & \textbf{2.89} \\
    \bottomrule[1.5pt]
  \end{tabular}
\end{table}

Interestingly, all models struggled more with quantifying changes in buildings than in roads, as evidenced by higher MAE and RMSE values for buildings in Table \ref{tab:comparison_quantification}). We attribute this to the greater complexity of buildings, whose architectural diversity poses a significant challenge for accurate quantification, in contrast to the relatively uniform and linear nature of road structures.

Despite these challenges, DeltaVLM consistently outperforms all baseline models, maintaining a clear advantage in both categories. This superior performance underscores the strength of our model in addressing the complexities of diverse change quantification tasks.

\subsubsection{Change Localization} We further evaluated DeltaVLM's ability to pinpoint the location of changes, focusing on roads and buildings, with results presented in Tables~\ref{tab:comparison_localization}. As shown in Table~\ref{tab:comparison_localization}, DeltaVLM outperforms all large VLMs. Compared to the best-performing baseline, Gemini-1.5-Pro, DeltaVLM achieves a substantial F1-score improvement of $26.2\%$, alongside significant gains in Precision and Recall of $26.62\%$ and $25.77\%$, respectively. The results highlight the challenges faced by models lacking dedicated change detection mechanisms in this task. For building localization, despite the increased complexity posed by urban structures, our model consistently maintains and even amplifies its performance advantage across all metrics. 

\begin{table}[t]
    \centering
    \begin{threeparttable}
      \centering
      \caption{Results on Change Localization for Roads and Buildings}
      \label{tab:comparison_localization}
      \setlength{\tabcolsep}{3.8pt}
      \renewcommand{\arraystretch}{1.1}
      \begin{tabular}{llccccc}
        \toprule[1.5pt]
        Category & Method & Prec.\tnote{1} & Rec.\tnote{2} & F1\tnote{3} & J. Sim.\tnote{4} & S. Acc.\tnote{5} \\
        \midrule\midrule
        \multirow{5}{*}{Roads} 
        & GPT-4o\cite{hurst2024gpt}       & 30.44 & 27.01 & 28.62 & 7.80  & 33.85 \\
        & Qwen-VL-Plus\cite{wang2024qwen2}  & 15.42 & 1.40  & 2.56  & 0.25  & 67.19 \\
        & GLM-4V-Plus\cite{glm2024chatglm}    & 21.99 & 33.32 & 26.49 & 7.93  & 6.79  \\
        & Gemini-1.5-Pro\cite{Reid2024Gemini1U} & 43.01 & 40.55 & 41.74 & 9.62  & 48.63 \\
        & \textbf{Ours}                        & \textbf{69.63} & \textbf{66.32} & \textbf{67.94} & \textbf{14.00} & \textbf{70.92} \\
        \midrule
        \multirow{5}{*}{Buildings} 
        & GPT-4o\cite{hurst2024gpt}        & 55.63 & 33.70 & 41.98 & 14.09  & 41.47  \\
        & Qwen-VL-Plus\cite{wang2024qwen2}  & 22.23 & 20.78  & 21.48  & 6.52  & 7.26 \\
        & GLM-4V-Plus\cite{glm2024chatglm}    & 38.98 & 57.83 & 46.57 & 17.93  & 17.11 \\
        & Gemini-1.5-Pro\cite{Reid2024Gemini1U} & 65.71 & 51.75 & 57.90 & 18.62  & 45.62 \\
        & \textbf{Ours}                        &\textbf{77.79} & \textbf{80.22} & \textbf{78.99} & \textbf{23.15} & \textbf{65.53} \\
        \bottomrule[1.5pt]
      \end{tabular}
      \begin{tablenotes}[para,flushleft]
            \footnotesize
            \item[1] Precision\,(\%). 
            \item[2] Recall\,(\%). 
            \item[3] F1-Score\,(\%).
            \item[4] Jaccard similarity\,(\%).
            \item[5] Subset accuracy\,(\%). 
          \end{tablenotes}
    \end{threeparttable}
\end{table}


Across both road and building localization tasks, DeltaVLM’s consistent and substantial performance improvements over general-purpose large VLMs in change localization highlight the effectiveness of our tailored architecture and domain-specific training in addressing the complex spatial requirements of RS imagery.

\subsubsection{Open-Ended Question Answering} To evaluate DeltaVLM's semantic understanding beyond change captioning, we evaluated its performance on open-ended question answering.  This task is more challenging than change captioning, as it requires the model to infer and respond to diverse user-specific queries. As shown in Table~\ref{tab:comparison_gpt_assist}, while GLM-4V-Plus performs competitively on METEOR and ROUGE-L, it lags behind on CIDEr, indicating weaker alignment with human-written answers. GPT-4o and Gemini-1.5-Pro achieve moderate performance, whereas Qwen-VL-Plus performs worst overall—especially on CIDEr—highlighting its limitations in processing complex multi-modal RS inputs. These results demonstrate DeltaVLM’s strong ability to understand and respond to diverse, open-ended instructions in the RSICA task.

\begin{table}[t]
    \centering
    \begin{threeparttable}
        \caption{Open-ended QA Results.}
        \label{tab:comparison_gpt_assist}
        \captionsetup{width=0.95\linewidth}
        \setlength{\tabcolsep}{4pt}
        \renewcommand{\arraystretch}{1.1}
        \begin{tabular}{l@{\hspace{0.2cm}}c@{\hspace{0.2cm}}c@{\hspace{0.2cm}}c@{\hspace{0.2cm}}c@{\hspace{0.2cm}}c@{\hspace{0.2cm}}c@{\hspace{0.2cm}}c}
            \toprule[1.5pt]
            Method & B-1\tnote{1} & B-2\tnote{1} & B-3\tnote{1} & B-4\tnote{1} & MTR\tnote{2} & R-L\tnote{3} & C-D\tnote{4} \\
            \midrule\midrule
            GPT-4o\cite{hurst2024gpt}       & 33.08 & 21.08 & 14.06 & 9.68  & 22.24 & 35.53 & 72.58 \\
            Qwen-VL-Plus\cite{wang2024qwen2}  & 24.75 & 12.55 & 6.70  & 3.88  & 16.69 & 27.74 & 27.22 \\
            GLM-4V-Plus\cite{glm2024chatglm}    & 34.27 & 22.38 & 15.66 & 11.43 & \textbf{22.48} &  \textbf{37.11} & 100.66 \\
            Gemini-1.5-Pro\cite{Reid2024Gemini1U} & 32.90 & 20.44 & 13.38 & 9.06  & 21.85 & 35.19 & 68.64 \\
            \textbf{Ours}                               & \textbf{36.67} & \textbf{27.09} & \textbf{20.62} & \textbf{16.21} &17.85 & 32.60 & \textbf{127.38} \\
            \bottomrule[1.5pt]
        \end{tabular}
      \begin{tablenotes}[para,flushleft]
        \scriptsize
        \item[1] BLEU-1/2/3/4. \item[2] METEOR. \item[3] ROUGE-L. \item[4] CIDEr-D.
      \end{tablenotes}
    \end{threeparttable}
\end{table}

\subsection{Ablation Study}

\begin{figure*}[htbp]
    \centering
    \includegraphics[width=\textwidth]{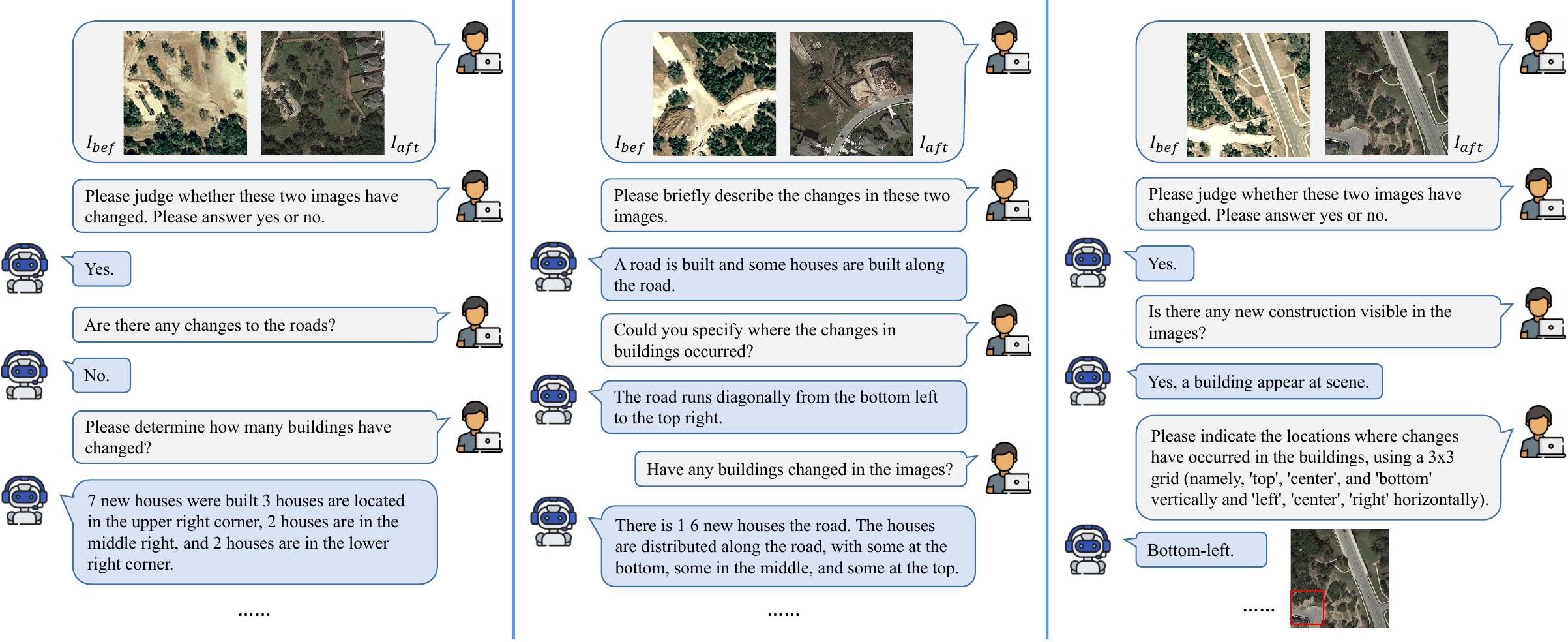}
    \caption{Demonstration of multi-round dialogue capability of DeltaVLM.}
    \label{fig:dialogue_examples}
\end{figure*}

To validate our proposed methods of Bi-VE fine-tuning and the cross-semantic relation measuring mechanism on the performance of remote sensing change analysis tasks, we conducted ablation studies on both change captioning and binary change classification tasks.

Table \ref{tab:comparison_captioning} compares DeltaVLM's performance under different conditions: without Bi-VE fine-tuning (denoted as ``w/o BiVE FT''), and without the CSRM (denoted as ``w/o CSRM''). The model without the CSRM mechanism shows a significant decline in various metrics. After freezing the parameters of the Bi-VE, the model exhibits improvements in all metrics. However, it still falls short compared to the complete DeltaVLM model. These experimental results indicate that the CSRM module is crucial for DeltaVLM, as its absence makes it difficult for the model to recognize visual difference information. Additionally, fine-tuning the parameters of the last two layers of the visual encoder indeed enhances its adaptability to RS scenarios, improving the model's capability to change captioning.

\begin{table}[t]
    \centering
    \begin{threeparttable}
        \setlength{\tabcolsep}{4pt}
        \renewcommand{\arraystretch}{1.1}
        \caption{Ablation Analysis on the Change Captioning Task}
        \label{tab:ablation_captioning}
            \begin{tabular}{lccccccc}
                \toprule[1.5pt]
                Method & B-1\tnote{1} & B-2\tnote{1} & B-3\tnote{1} & B-4\tnote{1} & MTR\tnote{2} & R-L\tnote{3} & C-D\tnote{4} \\
                \midrule\midrule
                w/o CSRM & 64.42 & 56.52 & 53.08 & 51.40 & 29.31 & 60.54 & 101.92 \\
                w/o Bi-VE FT & 84.24 & 75.62 & 67.91 & 61.40 & 39.29 & 74.73 & 134.76 \\
                DeltaVLM & \textbf{85.78} & \textbf{77.15} & \textbf{69.24} & \textbf{62.51} & \textbf{39.47} & \textbf{75.01} & \textbf{136.72} \\
                \bottomrule[1.5pt]
            \end{tabular}
        \begin{tablenotes}[para,flushleft]
            \scriptsize
            \item[1] BLEU-1/2/3/4. \item[2] METEOR. \item[3] ROUGE-L. \item[4] CIDEr-D.
        \end{tablenotes}
    \end{threeparttable}
\end{table}


In the binary change classification task, fine-tuning the Bi-VE, compared to the w/o Bi-VE FT condition led to a considerable improvement in the F1-score. Without the CSRM module again resulted in poor performance across all evaluation metrics. However, the model exhibited a strong bias toward predicting the ``no change'' class, indicating that it failed to detect meaningful differences without the semantic filtering provided by CSRM.

\begin{table}[t]
    \centering
    \captionsetup{width=0.9\linewidth}
    \setlength{\tabcolsep}{5pt}
    \renewcommand{\arraystretch}{1.1}
    \caption{Ablation Analysis on the Binary Change Classification Task}
    \label{tab:ablation_classification}
    \begin{tabular}{lcccc}
        \toprule[1.5pt]
        Method & Accuracy (\%) & Precision (\%) & Recall (\%) &F1 (\%) \\
        \midrule\midrule
        w/o CSRM & 50.13 & 75.00 & 0.31 & 0.62 \\
        w/o Bi-VE FT & 90.57 & \textbf{99.49} & 81.54 & 89.62 \\
        DeltaVLM & \textbf{93.99} & 96.29 & \textbf{91.49} & \textbf{93.83} \\
        \bottomrule[1.5pt]
    \end{tabular}
\end{table}

\subsection{Qualitative Results}

To further demonstrate DeltaVLM’s interactive capabilities, we evaluated its performance in multi-turn dialogue settings. Fig.\,\ref{fig:dialogue_examples} showcases example interactions where the model responds to a sequence of user queries based on bi-temporal RSIs.

These examples illustrate DeltaVLM’s ability to maintain conversational context, analyze image content, and generate coherent responses across diverse query types, including change detection, descriptive captioning, quantification, and localization. The model’s consistent and context-aware replies highlight its potential for for practical use in multi-turn dialogue scenarios for remote sensing change analysis.

\section{conclusion}

In this paper, we introduce interactive RSICA, a new task that redefines RS image analysis by enabling user-centric, query-driven exploration of temporal changes. To address this novel challenge, we propose DeltaVLM, an innovative end-to-end framework that integrates a selectively fine-tuned bi-temporal vision encoder, an IDPM with a CSRM mechanism, an instruction-guided Q-former for cross-modal alignment, and an LLM decoder for generating context-aware responses. To support this task, we constructed a large-scale dataset, ChangeChat-105k. Extensive experiments on this dataset demonstrate DeltaVLM’s superior performance across diverse interactive RSICA subtasks. Compared to existing multimodal VLMs, DeltaVLM excels in interactive, complex scenarios. Ablation studies further confirmed the effectiveness of each component. In future work, we will investigate innovative architectures for unified multimodal outputs, while enhancing the model’s reasoning capabilities and improving response efficiency.

\bibliographystyle{IEEEtran2}
\bibliography{refs}

\end{document}

%% file: imgs/instruction_types.tex
\begin{figure}[tbp]
    \centering
    \small
    \begin{examplebox}        
        {\color{Blue}\textbf{Type 1: Change Captioning}} \\
        \hspace{0.5em}{\color{Yellow}\texttt{<Img> <Img>}} Please briefly describe the changes in these two images. 
        \vspace{0.3em}\\
        {\color{Blue}\textbf{Type 2: Binary Change Classification}} \\
        \hspace{0.5em}{\color{Yellow}\texttt{<Img> <Img>}} Please judge whether these two images have changed.
        \hspace{0.5em}Please answer \texttt{yes} or \texttt{no}. 
        \vspace{0.3em}\\
        {\color{Blue}\textbf{Type 3: Category-Specific Change Quantification}} \\
        \hspace{0.5em}{\color{Yellow}\texttt{<Img> <Img>}} Please determine how many roads and buildings have changed? 
        \vspace{0.3em}\\
        {\color{Blue}\textbf{Type 4: Change Localization}} \\
        \hspace{0.5em}{\color{Yellow}\texttt{<Img> <Img>}} Please indicate the locations where changes have occurred in the buildings and roads, using a \texttt{3$\times$3} grid. 
        \vspace{0.3em}\\
        {\color{Blue}\textbf{Type 5: Open-ended QA}} \\
        \hspace{0.5em}{\color{Yellow}\texttt{<Img> <Img>}} Are there any changes in roads or lanes? \\
        \hspace{0.5em}{\color{Yellow}\texttt{<Img> <Img>}} What new features has appeared in the images? \\
        \hspace{0.5em}{\color{Yellow}\texttt{<Img> <Img>}} Is there any new construction visible in the images? 
        \vspace{0.3em}\\
        {\color{Blue}\textbf{Type 6: Multi-turn Conversation}} \\
        \hspace{0.5em}{\color{Yellow}\texttt{<Img> <Img>}} \\
        \hspace{0.5em}\textbf{Q1:} Please judge whether these two images have changed. Please answer \texttt{yes} or \texttt{no}. \\
        \hspace{0.5em}\textbf{Q2:} If changes have occurred, count the number of road and building changes separately. \\
        \hspace{0.5em}\textbf{Q3:} Based on the above analysis, please describe the changes of these two images in detail.
    \end{examplebox}
    \caption{Instruction types and examples in the ChangeChat-105k Dataset. The dataset employs rule-based methods for structured tasks (Types 1–4) and ChatGPT-generated responses for open-ended reasoning tasks (Types 5–6).}
    \label{tab:instruction_types}
\end{figure}

%% file: imgs/data-pipeline.tex
\begin{figure*}[tbp]
    \centering
    \small
    \begin{minipage}{\textwidth}
        \begin{tcolorbox}[
            colback=myLightBlueGrey, 
            colframe=myDarkBlueGrey, 
            title=\textbf{(a) System Message}, 
            fonttitle=\bfseries, 
            boxsep=3pt, 
            arc=0pt, 
            outer arc=0pt, 
            boxrule=1pt, 
            left=5pt, right=5pt, top=3pt, bottom=3pt 
        ]
            \noindent\begin{minipage}[c]{0.03\linewidth} 
                \centering\Large\color{myDarkBlueGrey}\faRobot 
            \end{minipage}%
            \hspace{1em}
            \begin{minipage}[c]{0.94\linewidth} 
                \textit{You are an AI visual assistant examining two remote sensing images. The differences you observe are summarized in five sentences that describe the changes between the two images. \ldots [Omitted for brevity].}
            \end{minipage}
        \end{tcolorbox}
    \end{minipage}

    \vspace{0.1cm} 
    \begin{minipage}{\textwidth}
        \begin{tcolorbox}[
            colback=myLightBlueGrey,
            colframe=myDarkBlueGrey,
            title=\textbf{(b) Few-shot Seed Example},
            fonttitle=\bfseries,
            boxsep=3pt,
            arc=0pt,
            outer arc=0pt,
            boxrule=1pt,
            left=5pt,
            right=5pt,
            top=3pt,
            bottom=3pt
        ]
            \begin{minipage}[t]{0.485\textwidth}
                \begin{tcolorbox}[
                    colback=white,
                    colframe=myMediumBlueGrey,
                    title=\textbf{Input Context},
                    boxsep=3pt,
                    arc=0pt,
                    outer arc=0pt,
                    boxrule=0.5pt,
                    left=5pt,
                    right=5pt,
                    top=3pt,
                    bottom=3pt,
                    equal height group=seed_example_boxes
                ]
                    \textbf{\color{Blue}Change Captions:}
                    \begin{itemize}[leftmargin=*,itemsep=0pt,parsep=0pt]
                        \item a row of buildings are constructed at the bottom.
                        \item a road with rows of villas built along replaces the trees in the bottom.
                        \vspace*{-1mm}
                        \item ...
                    \end{itemize}
                    
                    \textbf{\color{Blue}Change Counts:}\\
                    \texttt{\{"road": 1, "building": 10\}}
                    
                    \vspace{1mm}
                    \textbf{\color{Blue}Change Contours:}\\
                    \texttt{\{"road": [[[1.0, 0.84], ...]], "building": [[[0.95, 0.99], ...]]\}}
                \end{tcolorbox}
            \end{minipage}
            \hfill
            \begin{minipage}[t]{0.485\textwidth}
                \begin{tcolorbox}[
                    colback=white,
                    colframe=myMediumBlueGrey,
                    title=\textbf{Expected Output},
                    boxsep=3pt,
                    arc=0pt,
                    outer arc=0pt,
                    boxrule=0.5pt,
                    left=5pt,
                    right=5pt,
                    top=3pt,
                    bottom=3pt,
                    equal height group=seed_example_boxes
                    ]
                    \textbf{Question:}
                    Please determine \secondarykeyword{how many} roads and buildings have changed?
                    
                    \vspace{1mm}
                    \textbf{Answer:}
                    There is \keyword{1 new road} and \keyword{10 new houses}. \keyword{8 houses} above the road can be clearly seen, and the \keyword{other two} are located in the \secondarykeyword{lower left corner} and \secondarykeyword{below the road} and can only be seen a little bit.
                    
                    \vspace{1.5mm}
                    \small
                    \fcolorbox{mySlateGray!30}{white}{
                        \begin{minipage}{0.95\textwidth}
                            \textcolor{mySlateGray}{\textbf{Note:} \textit{This example demonstrates how ChatGPT performs in-context learning for change analysis tasks, adapting to the provided format.}}
                        \end{minipage}
                    }
                \end{tcolorbox}
            \end{minipage}
        \end{tcolorbox}
    \end{minipage}
    
    \vspace{0.1cm} 

    \begin{minipage}{\textwidth}
        \begin{tcolorbox}[
            colback=myLightBlueGrey, 
            colframe=myDarkBlueGrey, 
            title=\textbf{(c) Examples of Generated Conversational Data}, 
            fonttitle=\bfseries, 
            boxsep=3pt, arc=0pt, outer arc=0pt, boxrule=1pt, 
            left=5pt, right=5pt, top=3pt, bottom=3pt 
        ]
            \textbf{\color{Blue}Type I: General QA from Change Captions}
            \begin{tcolorbox}[
                colback=white, 
                sharp corners, 
                boxrule=0.5pt, 
                colframe=myFaintBlueGrey, 
                boxsep=1pt, left=3pt, right=3pt, top=1pt, bottom=1pt 
            ]
                \textbf{Question:} What is the \secondarykeyword{main change} that occurred in the area? \\
                \textbf{Answer:} A \keyword{new road was built}, and \keyword{several houses} were constructed \secondarykeyword{along it}, replacing what was previously a \secondarykeyword{wooded area}.
            \end{tcolorbox}

            \vspace{1mm}
            \textbf{\color{Blue}Type II: Fine-grained QA using Contour \& Count Info}
            \begin{tcolorbox}[
                colback=white, 
                sharp corners, 
                boxrule=0.5pt, 
                colframe=myFaintBlueGrey, 
                boxsep=1pt, left=3pt, right=3pt, top=1pt, bottom=1pt 
            ]
                \textbf{Question:} Can you \secondarykeyword{confirm the number} of new buildings and \secondarykeyword{describe their general location} based on the provided data? \\
                \textbf{Answer:} Yes, exactly \keyword{10 new buildings} were added. Based on the \keyword{contour data}, they are arranged in a row primarily along the \keyword{bottom part of the image}, adjacent to the \keyword{newly constructed road}.
            \end{tcolorbox}
        \end{tcolorbox}
    \end{minipage}
    \caption{Overview of the GPT-based data generation for open-ended QA. \textbf{(a)} The system message that defines the AI's role and task. \textbf{(b)} A few-shot seed example, showing the structured input (captions, counts, contours) and the desired output format. \textbf{(c)} Examples of the two main types of conversational data generated by our method.}
\label{fig:data_generation_pipeline}
\end{figure*}

%% file: paper.bbl
\begin{thebibliography}{10}
\providecommand{\url}[1]{#1}
\csname url@samestyle\endcsname
\providecommand{\newblock}{\relax}
\providecommand{\bibinfo}[2]{#2}
\providecommand{\BIBentrySTDinterwordspacing}{\spaceskip=0pt\relax}
\providecommand{\BIBentryALTinterwordstretchfactor}{4}
\providecommand{\BIBentryALTinterwordspacing}{\spaceskip=\fontdimen2\font plus
\BIBentryALTinterwordstretchfactor\fontdimen3\font minus \fontdimen4\font\relax}
\providecommand{\BIBforeignlanguage}[2]{{%
\expandafter\ifx\csname l@#1\endcsname\relax
\typeout{** WARNING: IEEEtran.bst: No hyphenation pattern has been}%
\typeout{** loaded for the language `#1'. Using the pattern for}%
\typeout{** the default language instead.}%
\else
\language=\csname l@#1\endcsname
\fi
#2}}
\providecommand{\BIBdecl}{\relax}
\BIBdecl

\bibitem{van2000remote}
C.~Van~Westen, ``Remote sensing for natural disaster management,'' \emph{Int. Arch. Photogramm. Remote Sens. Spat. Inf. Sci.}, vol.~33, no. B7/4; PART 7, pp. 1609--1617, 2000.

\bibitem{chowdhury2006driving}
R.~R. Chowdhury, ``Driving forces of tropical deforestation: The role of remote sensing and spatial models,'' \emph{Singap. J. Trop. Geogr.}, vol.~27, no.~1, pp. 82--101, 2006.

\bibitem{navalgund2007remote}
R.~R. Navalgund, V.~Jayaraman, and P.~Roy, ``Remote sensing applications: An overview,'' \emph{Curr. Sci.}, pp. 1747--1766, 2007.

\bibitem{bannari1995theoretical}
A.~Bannari, D.~Morin, G.~B{\'e}ni{\'e}, and F.~Bonn, ``A theoretical review of different mathematical models of geometric corrections applied to remote sensing images,'' \emph{Remote sensing reviews}, vol.~13, no. 1-2, pp. 27--47, 1995.

\bibitem{ding2025survey}
L.~Ding, D.~Hong, M.~Zhao, H.~Chen, C.~Li, J.~Deng, N.~Yokoya, L.~Bruzzone, and J.~Chanussot, ``A survey of sample-efficient deep learning for change detection in remote sensing: Tasks, strategies, and challenges,'' \emph{IEEE Geosci. Remote Sens. Mag.}, pp. 2--27, 2025.

\bibitem{qu2016deep}
B.~Qu, X.~Li, D.~Tao, and X.~Lu, ``Deep semantic understanding of high resolution remote sensing image,'' in \emph{2016 Int. Conf. Comput. Inf. Telecommun. Syst. (CITS)}.\hskip 1em plus 0.5em minus 0.4em\relax IEEE, 2016, pp. 1--5.

\bibitem{lobry2020rsvqa}
S.~Lobry, D.~Marcos, J.~Murray, and D.~Tuia, ``{RSVQA: V}isual question answering for remote sensing data,'' \emph{IEEE Trans. Geosci. Remote Sens.}, vol.~58, no.~12, pp. 8555--8566, 2020.

\bibitem{liu2022remote}
C.~Liu, R.~Zhao, H.~Chen, Z.~Zou, and Z.~Shi, ``Remote sensing image change captioning with dual-branch transformers: A new method and a large scale dataset,'' \emph{IEEE Trans. Geosci. Remote Sens.}, vol.~60, pp. 1--20, 2022.

\bibitem{radford2019language}
A.~Radford, J.~Wu, R.~Child, D.~Luan, D.~Amodei, I.~Sutskever \emph{et~al.}, ``Language models are unsupervised multitask learners,'' \emph{OpenAI blog}, vol.~1, no.~8, p.~9, 2019.

\bibitem{Lu2019ViLBERTPT}
J.~Lu, D.~Batra, D.~Parikh, and S.~Lee, ``{ViLBERT: P}retraining task-agnostic visiolinguistic representations for vision-and-language tasks,'' in \emph{Proc. Adv. Neural Inf. Process. Syst.}, 2019.

\bibitem{wei2021finetuned}
J.~Wei, M.~Bosma, V.~Y. Zhao, K.~Guu, A.~W. Yu, B.~Lester, N.~Du, A.~M. Dai, and Q.~V. Le, ``Finetuned language models are zero-shot learners,'' \emph{arXiv preprint arXiv:2109.01652}, 2021.

\bibitem{hu2025rsgpt}
Y.~Hu, J.~Yuan, C.~Wen, X.~Lu, Y.~Liu, and X.~Li, ``{RSGPT: A} remote sensing vision language model and benchmark,'' \emph{ISPRS J. Photogramm. Remote Sens.}, vol. 224, pp. 272--286, 2025.

\bibitem{kuckreja2024geochat}
K.~Kuckreja, M.~S. Danish, M.~Naseer, A.~Das, S.~Khan, and F.~S. Khan, ``Geochat: Grounded large vision-language model for remote sensing,'' in \emph{Proc. IEEE Conf. Comput. Vis. Pattern Recognit.}, 2024, pp. 27\,831--27\,840.

\bibitem{bazi2024rs}
Y.~Bazi, L.~Bashmal, M.~M. Al~Rahhal, R.~Ricci, and F.~Melgani, ``{RS-llava: A} large vision-language model for joint captioning and question answering in remote sensing imagery,'' \emph{Remote Sens.}, vol.~16, no.~9, p. 1477, 2024.

\bibitem{liu2023visual}
H.~Liu, C.~Li, Q.~Wu, and Y.~J. Lee, ``Visual instruction tuning,'' \emph{Proc. Adv. Neural Inf. Process. Syst.}, vol.~36, pp. 34\,892--34\,916, 2023.

\bibitem{rasti2018noise}
B.~Rasti, P.~Scheunders, P.~Ghamisi, G.~Licciardi, and J.~Chanussot, ``Noise reduction in hyperspectral imagery: Overview and application,'' \emph{Remote Sens.}, vol.~10, no.~3, p. 482, 2018.

\bibitem{openai2022chatgpt}
\BIBentryALTinterwordspacing
OpenAI, ``{ChatGPT}: Optimizing language models for dialogue,'' 2022, accessed: 2025-05-19. [Online]. Available: \url{https://openai.com/blog/chatgpt}
\BIBentrySTDinterwordspacing

\bibitem{singh1989review}
A.~Singh, ``Review article digital change detection techniques using remotely-sensed data,'' \emph{Int. J. Remote Sens.}, vol.~10, no.~6, pp. 989--1003, 1989.

\bibitem{nelson1983detecting}
R.~F. Nelson, ``Detecting forest canopy change due to insect activity using landsat mss,'' \emph{Photogramm. Eng. Remote Sens.}, vol.~49, no.~9, pp. 1303--1314, 1983.

\bibitem{nielsen1998multivariate}
A.~A. Nielsen, K.~Conradsen, and J.~J. Simpson, ``Multivariate alteration detection (mad) and maf postprocessing in multispectral, bitemporal image data: New approaches to change detection studies,'' \emph{Remote Sens. Environ.}, vol.~64, no.~1, pp. 1--19, 1998.

\bibitem{serra2003post}
P.~Serra, X.~Pons, and D.~Sauri, ``Post-classification change detection with data from different sensors: some accuracy considerations,'' \emph{Int. J. Remote Sens.}, vol.~24, no.~16, pp. 3311--3340, 2003.

\bibitem{blaschke2010object}
T.~Blaschke, ``Object based image analysis for remote sensing,'' \emph{ISPRS J. Photogramm. Remote Sens.}, vol.~65, no.~1, pp. 2--16, 2010.

\bibitem{daudt2018fully}
R.~C. Daudt, B.~Le~Saux, and A.~Boulch, ``Fully convolutional siamese networks for change detection,'' in \emph{Proc. Int. Conf. Image Process.}\hskip 1em plus 0.5em minus 0.4em\relax IEEE, 2018, pp. 4063--4067.

\bibitem{ronneberger2015u}
O.~Ronneberger, P.~Fischer, and T.~Brox, ``U-net: Convolutional networks for biomedical image segmentation,'' in \emph{Int. Conf. Med. Image Comput. Comput.-Assist. Interv. (MICCAI)}.\hskip 1em plus 0.5em minus 0.4em\relax Springer, 2015, pp. 234--241.

\bibitem{chen2021remote}
H.~Chen, Z.~Qi, and Z.~Shi, ``Remote sensing image change detection with transformers,'' \emph{IEEE Trans. Geosci. Remote Sens.}, vol.~60, pp. 1--14, 2021.

\bibitem{bandara2022transformer}
W.~G.~C. Bandara and V.~M. Patel, ``A transformer-based siamese network for change detection,'' in \emph{Proc. IEEE Int. Geosci. Remote Sens. Symp.}\hskip 1em plus 0.5em minus 0.4em\relax IEEE, 2022, pp. 207--210.

\bibitem{chen2020simple}
T.~Chen, S.~Kornblith, M.~Norouzi, and G.~Hinton, ``A simple framework for contrastive learning of visual representations,'' in \emph{Proc. Int. Conf. Mach. Learn.}\hskip 1em plus 0.5em minus 0.4em\relax PmLR, 2020, pp. 1597--1607.

\bibitem{bao2021beit}
H.~Bao, L.~Dong, S.~Piao, and F.~Wei, ``Beit: Bert pre-training of image transformers,'' \emph{arXiv preprint arXiv:2106.08254}, 2021.

\bibitem{palatucci2009zero}
M.~Palatucci, D.~Pomerleau, G.~E. Hinton, and T.~M. Mitchell, ``Zero-shot learning with semantic output codes,'' \emph{Proc. Adv. Neural Inf. Process. Syst.}, vol.~22, 2009.

\bibitem{longbotham2012multi}
N.~Longbotham, F.~Pacifici, T.~Glenn, A.~Zare, M.~Volpi, D.~Tuia, E.~Christophe, J.~Michel, J.~Inglada, J.~Chanussot \emph{et~al.}, ``Multi-modal change detection, application to the detection of flooded areas: Outcome of the 2009--2010 data fusion contest,'' \emph{IEEE J. Sel. Top. Appl. Earth Obs. Remote Sens.}, vol.~5, no.~1, pp. 331--342, 2012.

\bibitem{hoxha2022change}
G.~Hoxha, S.~Chouaf, F.~Melgani, and Y.~Smara, ``Change captioning: A new paradigm for multitemporal remote sensing image analysis,'' \emph{IEEE Trans. Geosci. Remote Sens.}, vol.~60, pp. 1--14, 2022.

\bibitem{you2016image}
Q.~You, H.~Jin, Z.~Wang, C.~Fang, and J.~Luo, ``Image captioning with semantic attention,'' in \emph{Proc. IEEE Conf. Comput. Vis. Pattern Recognit.}, 2016, pp. 4651--4659.

\bibitem{sun2024lightweight}
D.~Sun, Y.~Bao, J.~Liu, and X.~Cao, ``A lightweight sparse focus transformer for remote sensing image change captioning,'' \emph{IEEE J. Sel. Top. Appl. Earth Obs. Remote Sens.}, 2024.

\bibitem{liu2023progressive}
C.~Liu, J.~Yang, Z.~Qi, Z.~Zou, and Z.~Shi, ``Progressive scale-aware network for remote sensing image change captioning,'' in \emph{Proc. IEEE Int. Geosci. Remote Sens. Symp.}\hskip 1em plus 0.5em minus 0.4em\relax IEEE, 2023, pp. 6668--6671.

\bibitem{liu2023decoupling}
C.~Liu, R.~Zhao, J.~Chen, Z.~Qi, Z.~Zou, and Z.~Shi, ``A decoupling paradigm with prompt learning for remote sensing image change captioning,'' \emph{IEEE Transactions on Geoscience and Remote Sensing}, 2023.

\bibitem{Zhu2024SemanticCCBR}
Y.~Zhu, L.~Li, K.~Chen, C.~Liu, F.~Zhou, and Z.~X. Shi, ``Semantic-cc: Boosting remote sensing image change captioning via foundational knowledge and semantic guidance,'' \emph{IEEE Trans. Geosci. Remote Sens.}, vol.~62, pp. 1--16, 2024.

\bibitem{brown2020language}
T.~Brown, B.~Mann, N.~Ryder, M.~Subbiah, J.~D. Kaplan, P.~Dhariwal, A.~Neelakantan, P.~Shyam, G.~Sastry, A.~Askell \emph{et~al.}, ``Language models are few-shot learners,'' \emph{Proc. Adv. Neural Inf. Process. Syst.}, vol.~33, pp. 1877--1901, 2020.

\bibitem{noman2024cdchat}
M.~Noman, N.~Ahsan, M.~Naseer, H.~Cholakkal, R.~M. Anwer, S.~Khan, and F.~S. Khan, ``Cdchat: A large multimodal model for remote sensing change description,'' \emph{arXiv preprint arXiv:2409.16261}, 2024.

\bibitem{shi2021deeply}
Q.~Shi, M.~Liu, S.~Li, X.~Liu, F.~Wang, and L.~Zhang, ``A deeply supervised attention metric-based network and an open aerial image dataset for remote sensing change detection,'' \emph{IEEE Trans. Geosci. Remote Sens.}, vol.~60, pp. 1--16, 2021.

\bibitem{chen2020spatial}
H.~Chen and Z.~Shi, ``A spatial-temporal attention-based method and a new dataset for remote sensing image change detection,'' \emph{Remote Sens.}, vol.~12, no.~10, p. 1662, 2020.

\bibitem{antol2015vqa}
S.~Antol, A.~Agrawal, J.~Lu, M.~Mitchell, D.~Batra, C.~L. Zitnick, and D.~Parikh, ``Vqa: Visual question answering,'' in \emph{Proc. Int. Conf. Comput. Vis.}, 2015, pp. 2425--2433.

\bibitem{zhang2023spatial}
Z.~Zhang, L.~Jiao, L.~Li, X.~Liu, P.~Chen, F.~Liu, Y.~Li, and Z.~Guo, ``A spatial hierarchical reasoning network for remote sensing visual question answering,'' \emph{IEEE Trans. Geosci. Remote Sens.}, vol.~61, pp. 1--15, 2023.

\bibitem{wang2024earthvqa}
J.~Wang, Z.~Zheng, Z.~Chen, A.~Ma, and Y.~Zhong, ``Earthvqa: Towards queryable earth via relational reasoning-based remote sensing visual question answering,'' in \emph{Proc. AAAI Conf. Artif. Intell.}, vol.~38, no.~6, 2024, pp. 5481--5489.

\bibitem{chappuis2022prompt}
C.~Chappuis, V.~Zermatten, S.~Lobry, B.~Le~Saux, and D.~Tuia, ``Prompt-rsvqa: Prompting visual context to a language model for remote sensing visual question answering,'' in \emph{Proc. IEEE Conf. Comput. Vis. Pattern Recognit.}, 2022, pp. 1372--1381.

\bibitem{liu2024remote}
C.~Liu, J.~Zhang, K.~Chen, M.~Wang, Z.~Zou, and Z.~Shi, ``Remote sensing temporal vision-language models: A comprehensive survey,'' \emph{arXiv preprint arXiv:2412.02573}, 2024.

\bibitem{yuan2022change}
Z.~Yuan, L.~Mou, and X.~X. Zhu, ``Change-aware visual question answering,'' in \emph{Proc. IEEE Int. Geosci. Remote Sens. Symp.}\hskip 1em plus 0.5em minus 0.4em\relax IEEE, 2022, pp. 227--230.

\bibitem{li2022blip}
J.~Li, D.~Li, C.~Xiong, and S.~Hoi, ``Blip: Bootstrapping language-image pre-training for unified vision-language understanding and generation,'' in \emph{Proc. Int. Conf. Mach. Learn.}\hskip 1em plus 0.5em minus 0.4em\relax PMLR, 2022, pp. 12\,888--12\,900.

\bibitem{alayrac2022flamingo}
J.-B. Alayrac, J.~Donahue, P.~Luc, A.~Miech, I.~Barr, Y.~Hasson, K.~Lenc, A.~Mensch, K.~Millican, M.~Reynolds \emph{et~al.}, ``Flamingo: a visual language model for few-shot learning,'' \emph{Proc. Adv. Neural Inf. Process. Syst.}, vol.~35, pp. 23\,716--23\,736, 2022.

\bibitem{hurst2024gpt}
A.~Hurst, A.~Lerer, A.~P. Goucher, A.~Perelman, A.~Ramesh, A.~Clark, A.~Ostrow, A.~Welihinda, A.~Hayes, A.~Radford \emph{et~al.}, ``Gpt-4o system card,'' \emph{arXiv preprint arXiv:2410.21276}, 2024.

\bibitem{wang2024qwen2}
P.~Wang, S.~Bai, S.~Tan, S.~Wang, Z.~Fan, J.~Bai, K.~Chen, X.~Liu, J.~Wang, W.~Ge \emph{et~al.}, ``Qwen2-vl: Enhancing vision-language model's perception of the world at any resolution,'' \emph{arXiv preprint arXiv:2409.12191}, 2024.

\bibitem{glm2024chatglm}
T.~GLM, A.~Zeng, B.~Xu, B.~Wang, C.~Zhang, D.~Yin, D.~Zhang, D.~Rojas, G.~Feng, H.~Zhao \emph{et~al.}, ``Chatglm: A family of large language models from glm-130b to glm-4 all tools,'' \emph{arXiv preprint arXiv:2406.12793}, 2024.

\bibitem{Reid2024Gemini1U}
G.~Team, P.~Georgiev, V.~I. Lei, R.~Burnell, L.~Bai, A.~Gulati, G.~Tanzer, D.~Vincent, Z.~Pan, S.~Wang \emph{et~al.}, ``Gemini 1.5: Unlocking multimodal understanding across millions of tokens of context,'' \emph{arXiv preprint arXiv:2403.05530}, 2024.

\bibitem{zhang2024rs5m}
Z.~Zhang, T.~Zhao, Y.~Guo, and J.~Yin, ``{RS5M and GeoRSCLIP: A} large scale vision-language dataset and a large vision-language model for remote sensing,'' \emph{IEEE Trans. Geosci. Remote Sens.}, 2024.

\bibitem{dai2023instructblip}
W.~Dai, J.~Li, D.~Li, A.~Tiong, J.~Zhao, W.~Wang, B.~Li, P.~N. Fung, and S.~Hoi, ``Instructblip: Towards general-purpose vision-language models with instruction tuning,'' \emph{Proc. Adv. Neural Inf. Process. Syst.}, vol.~36, pp. 49\,250--49\,267, 2023.

\bibitem{zhan2025skyeyegpt}
Y.~Zhan, Z.~Xiong, and Y.~Yuan, ``Skyeyegpt: Unifying remote sensing vision-language tasks via instruction tuning with large language model,'' \emph{ISPRS J. Photogramm. Remote Sens.}, vol. 221, pp. 64--77, 2025.

\bibitem{deng2025changechat}
P.~Deng, W.~Zhou, and H.~Wu, ``Changechat: An interactive model for remote sensing change analysis via multimodal instruction tuning,'' in \emph{ICASSP 2025-2025 IEEE International Conference on Acoustics, Speech and Signal Processing (ICASSP)}.\hskip 1em plus 0.5em minus 0.4em\relax IEEE, 2025, pp. 1--5.

\bibitem{liu2024change}
C.~Liu, K.~Chen, H.~Zhang, Z.~Qi, Z.~Zou, and Z.~Shi, ``Change-agent: Towards interactive comprehensive remote sensing change interpretation and analysis,'' \emph{IEEE Trans. Geosci. Remote Sens.}, 2024.

\bibitem{fang2023eva}
Y.~Fang, W.~Wang, B.~Xie \emph{et~al.}, ``{EVA: E}xploring the limits of masked visual representation learning at scale,'' in \emph{Proc. IEEE Conf. Comput. Vis. Pattern Recognit.}, 2023, pp. 19\,358--19\,369.

\bibitem{Cho2014LearningPR}
K.~Cho, B.~van Merrienboer, Çaglar G{\"u}lçehre, D.~Bahdanau, F.~Bougares, H.~Schwenk, and Y.~Bengio, ``Learning phrase representations using rnn encoder–decoder for statistical machine translation,'' in \emph{Conf. Empir. Methods Nat. Lang. Process.}, 2014.

\bibitem{zheng2023judging}
L.~Zheng, W.-L. Chiang, Y.~Sheng, S.~Zhuang, Z.~Wu, Y.~Zhuang, Z.~Lin, Z.~Li, D.~Li, E.~Xing \emph{et~al.}, ``Judging llm-as-a-judge with mt-bench and chatbot arena,'' \emph{Proc. Adv. Neural Inf. Process. Syst.}, vol.~36, pp. 46\,595--46\,623, 2023.

\bibitem{touvron2023llama}
H.~Touvron, T.~Lavril, G.~Izacard, X.~Martinet, M.-A. Lachaux, T.~Lacroix, B.~Rozi{\`e}re, N.~Goyal, E.~Hambro, F.~Azhar \emph{et~al.}, ``Llama: Open and efficient foundation language models,'' \emph{arXiv preprint arXiv:2302.13971}, 2023.

\bibitem{loshchilov2017decoupled}
I.~Loshchilov and F.~Hutter, ``Decoupled weight decay regularization,'' \emph{arXiv preprint arXiv:1711.05101}, 2017.

\bibitem{papineni2002bleu}
K.~Papineni, S.~Roukos, T.~Ward, and W.-J. Zhu, ``{BLEU}: a method for automatic evaluation of machine translation,'' in \emph{Proc. 40th Annu. Meet. Assoc. Comput. Linguist.}, 2002, pp. 311--318.

\bibitem{banerjee2005meteor}
S.~Banerjee and A.~Lavie, ``{METEOR}: An automatic metric for mt evaluation with improved correlation with human judgments,'' in \emph{Proc. ACL Workshop Intrinsic Extrinsic Eval. Mach. Transl. Summ.}, 2005, pp. 65--72.

\bibitem{lin2004rouge}
C.-Y. Lin, ``{Rouge: A} package for automatic evaluation of summaries,'' in \emph{Proc. Workshop Text Summ. Branches Out}, 2004, pp. 74--81.

\bibitem{Vedantam2014CIDErCI}
R.~Vedantam, C.~L. Zitnick, and D.~Parikh, ``{CIDEr: C}onsensus-based image description evaluation,'' \emph{Proc. IEEE Conf. Comput. Vis. Pattern Recognit.}, pp. 4566--4575, 2014.

\end{thebibliography}
